\documentclass[a4paper,10pt]{llncs}
\usepackage[utf8]{inputenc}
\usepackage{amsmath, amssymb}
\usepackage{graphicx}
\usepackage{xspace} 
\usepackage{tikz}
\usetikzlibrary{calc}
\usepackage{todonotes}
\usepackage[defblank]{paralist}
\usepackage{url}
\usepackage{times}
\usepackage[hidelinks]{hyperref}
\usetikzlibrary{petri}

\title{Conformance Checking with Uncertainty\\ via SMT \\ (Extended Version)}
\titlerunning{Conformance Checking with Uncertainty via SMT (Extended Version)}
\author{Paolo Felli\inst{1} \and
Alessandro Gianola\inst{1} \and
Marco Montali\inst{1} \and \\
 Andrey Rivkin\inst{1} \and Sarah Winkler\inst{1}}
\authorrunning{Felli, Gianola, Montali, Rivkin, Winkler}
\institute{%
 Free University of Bozen-Bolzano, Bolzano, Italy
\email{\{pfelli,gianola,montali,rivkin,winkler\}@inf.unibz.it}
}

\newcommand{\set}[1]{\{#1\}} 
\newcommand{\seq}[2][n]{\tup{{#2_1},\dots,{#2_{#1}}}}
\newcommand{\sset}[2][n]{\set{{#2_1},\dots,{#2_{#1}}}}
\renewcommand{\vec}[1]{\mathbf{#1}} 
\newcommand{\pre}[1]{{{}^\bullet{#1}}} 
\newcommand{\post}[1]{{{#1}^\bullet}} 
\newcommand{\NN}{\mathcal N} 
\newcommand{\goto}[1]{\mathrel{\raisebox{-2pt}{$\xrightarrow{#1}$}}} 
\renewcommand{\empty}{{\gg}}
\newcommand{\tup}[1]{\langle #1 \rangle}
\newcommand{\bool}{\mathtt{bool}}
\newcommand{\integer}{\mathtt{int}}
\newcommand{\str}{\mathtt{string}}
\newcommand{\rational}{\mathtt{rat}}

\newcommand{\E}{\mathcal{E}}		
\newcommand{\restr}[2]{\left.#1\right|_{#2}} 

\newcommand{\tool}{\texttt{cocomot}\xspace}

\newcommand{\LQA}{\ensuremath{\mathcal{LQA}}\xspace}

\newcommand{\LIA}{\ensuremath{\mathcal{LIA}}\xspace}

\newcommand{\transvar}{\mathtt{S}} 
\newcommand{\markvar}{\mathtt{M}} 
\newcommand{\datavar}{\mathtt{X}} 
\newcommand{\dropvar}{\mathtt{drop}} 
\newcommand{\actvar}{\mathtt{A}} 
\newcommand{\tdvar}{\mathtt{D}} 
\newcommand{\tsvar}{\mathtt{T}} 
\newcommand{\posvar}{\mathtt{P}} 
\newcommand{\nthvar}{\mathtt{L}} 
\newcommand{\distvar}{\mathtt{d}} 

\newcommand{\ite}{\mathit{ite}} 
\newcommand{\conf}[1]{#1.\mathit{conf}} 

\newcommand{\constraint}{c}
\newcommand{\runsof}[1]{Runs(#1)}
\newcommand{\firingsof}[1]{\mathcal{F}(#1)}
\newcommand{\moves}{\mathit{Moves}_{\NN}}
\newcommand{\alignment}[1]{Align(\NN,{#1})}

\newcommand{\eventsplus}{\E^{\empty}}
\newcommand{\firingsplus}{\mathcal{F}^{\empty}}

\newcommand{\procrun}{\vec f}
\newcommand{\logtrace}{\vec e}

\newcommand{\syncmove}[4]{\tikz[baseline=-0.5ex]{\node[scale=.7, inner sep=0pt]{$%
\begin{array}{|@{\,}l@{\quad}l@{\,}|}%
\hline{#1}&{#2}\\\hline{#3}&{#4}\\\hline\end{array}$}}}

\newcommand{\modelmove}[2]{\tikz[baseline=-0.5ex]{\node[scale=.7, inner sep=0pt]{$%
\begin{array}{|@{\,}l@{\quad}l@{\,}|}%
\hline\multicolumn{2}{|c|}{\empty}\\\hline{#1}&{#2}\\\hline\end{array}$}}}

\newcommand{\m}[1]{\mathsf{#1}\xspace}

\newcommand{\longversion}[2]{#1} 

\newcommand{\ueventId}{\textsc{id}\xspace}
\newcommand{\ueventConfidence}{\mathit{conf}\xspace}
\newcommand{\ueventLabels}{\textsc{la}\xspace}
\newcommand{\ueventTss}{\textsc{ts}\xspace}
\newcommand{\ulogtrace}{\textbf{u}\logtrace\xspace}
\newcommand{\uevent}{\textit{ue}\xspace}
\newcommand{\comp}[2]{{#1}.{#2}}

\newcommand{\R}{\mathcal{R}}
\newcommand{\eid}[1]{\texttt{\#}_{#1}}
\newcommand{\lab}{\mathit{lab}}

\usepackage{comment}

\begin{document}
\maketitle

\begin{abstract}
Logs of real-life processes often feature uncertainty pertaining the recorded timestamps, data values, and/or events. We consider the problem of checking conformance of uncertain logs against data-aware reference processes. Specifically, we show how to solve it via SMT encodings, lifting previous work on data-aware SMT-based conformance checking to this more sophisticated setting. Our approach is modular, in that it homogeneously accommodates for different types of uncertainty. Moreover, using appropriate cost functions, different conformance checking tasks can be addressed. We show the correctness of our approach and witness feasibility through a proof-of-concept implementation.
\end{abstract}


\section{Introduction}
\label{sec:intro}

Process mining is a well-established field of research at the intersection between BPM and data science. The vast majority of process mining tasks assumes that their input event data provide an accurate and complete digital footprint of reality \cite{Aals11}. In many settings, this is an unrealistic assumption: events may be missing or totally/partially wrongly recorded, due to various factors such as human errors, faulty loggers, errors in the acquisition of events (e.g., through sensors), etc. To mitigate this issue, two lines of research emerged lately. The first deals with methodologies and techniques to improve the quality of event data, thus handling uncertainty in the data preparation phase \cite{WynS19}. The second  aims instead at incorporating the management of uncertainty within the process mining tasks themselves, leading to a new generation of process mining techniques where process models \cite{LABP21,PolK22,BMMP21,AMMP22} and/or event logs \cite{PegoraroUA21,CMMF18} explicitly address different kinds of uncertainty. 

Surprisingly enough, the latter has received much less attention from the community. In this work, we aim at contributing to the advancement of process mining on uncertain data, considering in particular the problem of conformance checking \cite{CDSW18}. Specifically, our contribution is twofold:
\begin{compactenum}
\item We introduce a framework for data-aware conformance checking over uncertain logs, through a suitably extended notion of alignment. The framework employs Data Petri nets \cite{Mannhardt18} for reference process models, and addresses event logs incorporating sophisticated forms of uncertainty, pertaining the recorded timestamps, data values, and/or events. Notably, the framework comes with a generic cost function whose components can be flexibly instantiated to homogeneously account for a variety of measures required for computing optimal alignments.
\item We devise a corresponding operational counterpart to effectively attack the problem of computing alignments and their costs. Instead of relying on ad-hoc algorithmic techniques, our approach builds on and extends \cite{cocomot} to encode the problem into the well-established automated reasoning framework of SMT. This allows us to employ state-of-the-art SMT solvers. 
\end{compactenum}

To handle uncertainty in the log, we follow the approach in~\cite{PegoraroUA21}, where the log is explicitly enriched with annotations reflecting the degree and nature of uncertainty. Such annotations may be derived from operational characteristics of the information system recording the event data (considering its logging precision and reliability), and/or by directly attaching them to the generated events. For instance, the log may be enriched with explicit details on the coarseness or precision of an automatic logging device (such as a sensor); alternatively, uncertainty-related annotations may be derived from domain knowledge on the precision and frequency of a specific human activity.
In particular, our framework accounts for four main types of uncertain event data.
\begin{compactitem}[$\bullet$]
\item \emph{Uncertain events}: these are recorded in a log trace but come with a known \emph{confidence value}, capturing the degree of (un)certainty about the fact that a recorded event actually happened at all during the process execution.  
\item \emph{Uncertain timestamps}: due to coarseness of the logging activity, events are in general not totally ordered, but come with a fixed range of possible timestamp values. This calls for considering multiple possible orderings and treating  a log trace as \emph{a set} of events rather than a sequence.
\item \emph{Uncertain activities}: this pertains events whose reference activity is not certainly known. Hence, the event comes with a candidate set of possible activities (each with its own confidence value).
\item \emph{Uncertain data values}: in the execution of data-aware processes, for instance due to sensor precision, event data attributes may come with both coarseness and ambiguity. Specifically, the log may only record a set of possible values or an interval for a given attribute, requiring all possible values to be considered. 
\end{compactitem} 
%
We stress that the notion of confidence used here should not be confused with that of probability: 
it measures the degree of trust in the recorded behaviour, which has nothing to do with the likelihood/frequency of such a behaviour. 

To account for these different types of uncertain event data, we borrow from \cite{PegoraroUA21} and adapt to our data-aware setting the notion of \emph{realization}. A realization of a log trace with uncertainty is an \emph{ordered sequence} of events in which the uncertainty of all types of event data as above is resolved. Our task 
then concretely becomes as follows: given a Data Petri net and a log trace  with uncertainty, find some realization of that trace that admits an \emph{optimal alignment}, i.e., an alignment of minimal cost among all possible realizations for that log trace. Differently from \cite{PegoraroUA21}, the confidence values of the original trace are used as an essential component for measuring the cost incurred in selecting realizations. 

Crucially, since we are in a data-aware setting, a log trace may correspond to infinitely many possible realizations. This is handled symbolically thanks to our SMT-based approach. 

\smallskip
The rest of the paper is organized as follows. 
First, in Sec.~\ref{sec:preliminaries} we recall the required preliminaries. 
Then, in Sec.~\ref{sec:alignments} we fix the shape of traces in event logs with uncertainty and the notion of alignments.
In Sec.~\ref{sec:costmodel} we detail the cost components that must be considered in the setting with uncertain even data and that we use to define the conformance checking task. We discuss separately one main cost component: the notion of  data-aware alignment cost function (in Sec.~\ref{ssec:distance}). In Sec.~\ref{sec:encoding} we illustrate our SMT-based encoding and we report on the implementation.
We conclude in Sec.~\ref{sec:conclusion}. 

This paper is the extended version of a conference paper accepted at the 20th International Conference on Business Process Management (BPM 2022) \cite{bpm22}.



\section{Preliminaries}
\label{sec:preliminaries}

In this section we recall data Petri nets (DPNs) and their execution semantics, and the main notions of the machinery behind our approach, namely SMT. 

\subsection{Data Petri Nets}
\label{sec:dpn}
We use Data Petri nets (DPNs) for modelling multi-perspective processes, adopting the same formalization as in \cite{cocomot,Mannhardt18}.  
For lack of space, in what follows we only recall the definitions and notation required for our technical development, referring the reader to \cite{cocomot,Mannhardt18} for further details. 
 
Let $V$ be a set of \emph{process variables}, each with a type and an associated domain: booleans (type $\texttt{bool}$), integers (\texttt{int}), rationals (\texttt{rat}) or strings (\texttt{string}). 
We consider two disjoint sets of annotated variables $V^r = \{v^r \mid v\,{\in}\,V\}$ and $V^w = \{v^w \mid v\,{\in}\,V\}$ to be  read and written by process activities, as explained below.   
Based on these, we define constraints according to the grammar for $\constraint$: 
\begin{align*}
 c &::= v_b \mid b \mid n \geq n \mid r \geq r \mid r > r \mid s = s \mid c \wedge c \mid \neg c &
 s &::= v_s \mid t \\
 n &::= v_z \mid z \mid n + n \mid - n &
 r &::= v_r \mid q \mid r + r \mid - r
\end{align*}
where
$v_b \in V_{\bool}$, $b \in \mathbb B$,
$v_s \in V_{\str}$, $t \in \mathbb S$,
$v_z \in V_{\integer}$, $z \in \mathbb Z$,
$v_r \in V_{\rational}$, and $q \in \mathbb Q$. 
Standard equivalences apply, hence disjunction (i.e., $\lor$) and comparisons $>$, $\neq$, $<$, $\leq$ can be used as well ($\texttt{bool}$ and $\texttt{string}$ only support (in)equality).
The set of constraints over variables $V$ is denoted $\mathcal C(V)$.
These form the basis for expressing conditions on the values of variables that are read and written during the execution of process activities.  
For instance, a constraint $(v_1^r > v_2^r)$ dictates that the current value of variable $v_1$ is greater than the current value of $v_2$. Similarly, $(v_1^w > v_2^r + 1) \land (v_1^w < v_3^r)$ requires that the new value given to $v_1$ (i.e., assigned as a result of the execution of the activity to which this constraint is attached) is greater than the current value of $v_2$ plus $1$, and smaller than $v_3$. 

\begin{definition}[DPN]
A tuple $\NN = (P, T, F, \ell, A, V,guard)$ is a \emph{Petri net with data} (DPN), where:
\begin{compactitem}
\item $(P, T, F, \ell)$ is a Petri net with two non-empty disjoint sets of places $P$ and transitions $T$, 
a flow relation $F:(P \times T)\cup(T \times P)\rightarrow\mathbb{N}$ and a labeling function $\ell:T\to A\cup \set{\tau}$, where $A$ is a finite set of activity labels and $\tau$ is a special symbol denoting silent transitions;
\item $V$ is a set of typed process variables; and
\item $guard\colon T \to \mathcal C(V)$ is a guard assignment (for $t\in T$ with $\ell(t)=\tau$ we assume that $guard(t)$ does not use variables in $V^w$). 
\end{compactitem} 
\end{definition}
As customary, given $x\in P \cup T$, 
we use $\pre{x}:=\set{y\mid F(y,x)>0}$ to denote the \emph{preset} of $x$ and $\post{x}:=\set{y\mid F(x,y)>0}$ to denote the \emph{postset} of $x$. 

To assign values to variables, we consider a \emph{state variable assignment}, i.e., a total function $\alpha$ that assigns a value (of the right type) to each variable in $V$. 
A \emph{state} in a DPN $\NN$ is a pair $(M,\alpha)$ constituted by a marking $M\colon P\rightarrow\mathbb{N}$ for the underlying Petri net $(P, T, F, \ell)$, plus a state variable assignment $\alpha$. Therefore, a state simultaneously accounts for the control flow progress and for the current values of all variables in $V$, as specified by $\alpha$. 

Given $\NN$, we fix one state $(M_I,\alpha_0)$ as \emph{initial}, where $M_I$ is the initial marking of the underlying Petri net $(P, T, F, \ell)$ and $\alpha_0$ specifies the initial value of all variables in $V$. 
Similarly, we denote the final marking as $M_F$, and call \emph{final} any state of $\NN$ of the form $(M_F,\alpha_F)$ for some $\alpha_F$.

We now define when a Petri net transition may fire from a given state $(M, \alpha)$. 
Informally, a \emph{transition firing} is a couple $(t,\beta)$ where $t\in T$ and $\beta$ is a function used to determine the new values of variables after the transition has fired.
The step yields a new state $(M', \alpha')$, and is denoted $(M,\alpha)\goto{(t_n, \beta_n)}(M',\alpha')$. 
A transition firing is \emph{valid} in a state $(M,\alpha)$ when $t$ is enabled in $M$ and $\alpha$ satisfies the constraint associated to $t$. 
The formal definition can be found, e.g., in~\cite{cocomot,Mannhardt18}. 

Based on this single-step transition firing, we say that a state $(M',\alpha')$ is \emph{reachable} in a DPN with initial state $(M_I,\alpha_0)$ iff there exists a sequence of valid transition firings of the form 
$\procrun =\langle(t_1, \beta_1), \dots, (t_n, \beta_n)\rangle$ such that $(M_I,\alpha_0)\goto{(t_1, \beta_1)} \ldots\goto{(t_n, \beta_n)}(M',\alpha')$. 
Moreover, such a sequence $\procrun$ is called a \emph{process run} of $\NN$ if $(M_I,\alpha_0)\goto{\procrun} (M_F,\alpha_F)$ for some $\alpha_F$, i.e., if the run leads to a final state.  
As in~\cite{cocomot,MannhardtLRA16}, we restrict to DPNs 
where at least one final state is reachable.  

We denote the set of transition firings of a DPN $\NN$ by $\firingsof{\NN}$, and the set of process runs by $\runsof{\NN}$. 

\begin{example}
\label{exa:1}
Let $\NN$ be as shown (with initial marking $[p_0]$ and final marking $[p_3]$). 
$\runsof{\NN}$ contains, e.g.,
$\langle (\mathsf{a}, \{x^w\mapsto 2\}), (\mathsf{b}, \{y^w\mapsto 1\}), (\mathsf{c}, \{ x^r\mapsto 2, y^r\mapsto 1 \})\rangle$ and
$\langle (\mathsf{a}, \{x^w\mapsto 1\}), (\mathsf{b}, \{ y^w\mapsto 1 \}), (\mathsf{d}, \{y^r\mapsto 1, x^r \mapsto 1\})\rangle$, for $\alpha_0 = \{x,y\mapsto 0\}$. 
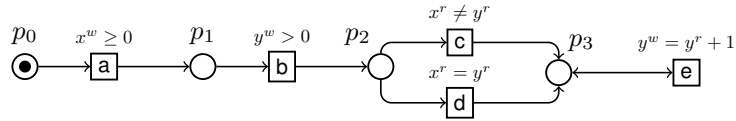
\begin{center}
\begin{tikzpicture}[node distance=13mm,font=\normalfont]
\tikzstyle{place}=[draw, circle, line width=.7pt]
\tikzstyle{trans}=[draw, rectangle, line width=.7pt, scale=.8, minimum height=4mm, minimum width=4mm, ]
\tikzstyle{goto}=[->, line width=.6pt]
\tikzstyle{tlabel}=[yshift=3.8mm, scale=.7,]
\node[place,tokens=1,label=above:$p_0$] (p0) {};
\node[trans, right of=p0] (a) {\textsf{a}};
\node[tlabel] at (a) {$x^w \geq 0$};
\node[place, right of=a, label=above:$p_1$] (p1) {};
\node[trans, right of=p1] (b) {\textsf{b}};
\node[tlabel] at (b) {$y^w > 0$};
\node[place, right of=b] (p2) {};
\node[xshift=-3mm, yshift=4mm] at (p2) {$p_2$};
\node[trans, right of=p2, yshift=4mm] (c) {\textsf{c}};
\node[tlabel] at (c) {$x^r \neq y^r$};
\node[place, right of=c,yshift=-4mm] (p3) {};
\node[xshift=3mm, yshift=4mm] at (p3) {$p_3$};
\node[trans, right of=p3, xshift=8mm] (e) {\textsf{e}};
\node[tlabel, above of=e, yshift=-13mm] {$y^w = y^r + 1$};
\node[trans, right of=p2, yshift=-6mm] (d) {\textsf{d}};
\node[tlabel, above of=d, yshift=-13mm] {$x^r = y^r$};
\draw[goto] (p0) -- (a);
\draw[goto, rounded corners] (p2) |- (d);
\draw[goto, rounded corners] (d) -| (p3);
\draw[goto, rounded corners] (p2) |- (c);
\draw[goto, rounded corners] (c) -| (p3);
\draw[goto] (a) -- (p1);
\draw[goto] (p1) -- (b);
\draw[goto] (b) -- (p2);
\draw[goto, <->] (p3) -- (e);
\end{tikzpicture}
\end{center}
\end{example}

\subsection{Satisfiability Modulo Theories (SMT)}
\label{sec:smt}
The classic propositional satisfiability (SAT) problem amounts to, given a propositional formula $\varphi$, 
either find an assignment $\nu$ under which $\varphi$ evaluates to true, or
detect that $\varphi$ is unsatisfiable.
E.g., given the formula $(p \vee q) \wedge (\neg p \vee r) \wedge (\neg r \vee \neg q)$, a satisfying assignment is $\nu(p) = \nu(r) = \top$, $\nu(q) = \bot$.
The SMT problem \cite{BarrettT18} is an extension of SAT that consists of establishing satisfiability of a formula $\varphi$ whose 
language enriches 
propositional formulas with constants and operators from one or more theories 
$\mathcal T$ (e.g., arithmetics, bit-vectors, arrays, uninterpreted functions).
In this paper, we only consider the theories of linear integer
and rational arithmetic ($\LIA$ and $\LQA$).
For instance, the SMT formula $a > 1 \wedge (a + b = 10 \vee a - b = 20) \wedge p$, where $a$, $b$ are integer and $p$ is a propositional variable, is satisfiable
by the assignment $\nu$ such that $\nu(a) = \nu(b) = 5$ and $\nu(p) = \top$.
Another important problem studied in the area of SMT and relevant to this paper is the one of Optimization Modulo Theories (OMT) \cite{SebastianiT15}. The OMT problem asks, given a formula $\varphi$, to
find a satisfying assignment of $\varphi$ that minimizes or maximizes a given objective
expression.
SMT-LIB  \cite{smt-lib} is an initiative aiming at providing an extensive on-line library of benchmarks and promoting the adoption of common languages and interfaces for SMT solvers. In this paper, we employ the SMT solvers Yices 2 \cite{Dutertre14} and Z3 \cite{deMouraB08}.

\section{Event Logs with Uncertainty and Alignments}
\label{sec:alignments}

Let $\mathit{ID}$ be a finite set of event identifiers, $A$ be a finite set of activity labels, and $\mathit{TS}$ be 
a totally ordered set of possible timestamps (for simplicity, we use $\mathbb{N}$).

\begin{definition}
An event with uncertainty is a tuple $\uevent = \tup{\ueventId,\ueventConfidence,\ueventLabels,\ueventTss,\alpha}$ s.t.  
\begin{compactitem}
	\item $\ueventId\in \mathit{ID}$ is an event identifier;
	\item $0< \ueventConfidence \leq 1$ expresses the confidence that the event actually happened. We say that the event is an uncertain event whenever $\ueventConfidence<1$;
	\item $\ueventLabels = \set{b_1\colon p_1, \ldots, b_n\colon p_n}$ is a finite, non-empty subset of activity labels $b_i\in A$, each associated to a confidence value $0 < p_i \leq 1$ so that $\sum_{i=1}^n p_i =1$; 
	\item $\ueventTss$ is either a finite set of timestamps in $\mathit{TS}$ or an interval over $\mathit{TS}$;
	\item with some abuse of notation, $\alpha$ is a (possibly partial) function returning for variables in $V$ a \emph{finite set} of values in the domain of $v$ or an interval over such domain (if $v$ is of type \normalfont{\texttt{int}} or \normalfont{\texttt{rat}}).
\end{compactitem}
\end{definition}

Given an event $\uevent = \tup{\ueventId,\ueventConfidence,\ueventLabels,\ueventTss,\alpha}$, we denote its components 
by $\ueventId(\uevent)$, $\ueventConfidence(\uevent)$, $\ueventLabels(\uevent)$, $\ueventTss(\uevent)$ and  $\alpha(\uevent)$, respectively.  

Note that we do not associate confidence values to timestamps, along the lines of \cite{PegoraroUA21}. 
We also do not consider timestamp values following any kind of distribution, e.g., a normal distribution, as this would make the encoding in Section~\ref{sec:encoding} computationally too challenging.

\begin{definition}
A log \emph{trace with uncertainty} $\ulogtrace$ is a finite set of events with uncertainty, such that all event identifiers are unique.
\end{definition}

Thus, there is no fixed order among the events in a trace with uncertainty.
An \emph{event log} $L$ is a multiset of 
log traces with uncertainty.

\begin{example}
\label{exa:2}
Consider $\NN$ from Ex. \ref{exa:1}. For simplicity, we use natural numbers for timestamps. The following are three possible traces with uncertainty: 
\begin{small}
\begin{align*}
\ulogtrace_1 =& \{
 \tup{\eid{1},.25, \set{\m a\,{:}\,1}, [0\text{-}5], \{x\,{\mapsto}\,\set{2,3}\}},\;
 \tup{\eid{2},.9, \set{\m b\,{:}\,.8, \m c\,{:}\,.2}, \{2\}, \{y\,{\mapsto}\,\set{1}\}}\\
\ulogtrace_2 =& \{
 \tup{\eid{3},1, \set{\m a\,{:}\,1}, \{0\}, \{x\,{\mapsto}\,[1,6.5]\}},\;
 \tup{\eid{4},1, \set{\m b\,{:}\,1}, \{2\}, \{y\,{\mapsto}\,\set{1}\}},\; \\
 &\phantom{\{} 
 \tup{\eid{5},1, \set{\m c\,{:}\,1}, \{3\}, \emptyset} \} \\
\ulogtrace_3 =& \{
 \tup{\eid{6},1, \set{\m a\,{:}\,1}, \{2\}, \{x\,{\mapsto}\,\set{6}\}},\;
 \tup{\eid{7},1, \set{\m b\,{:}\,1}, \{2\}, \{y\,{\mapsto}\,\set{1}}\}
\}
\end{align*}
\end{small}
For instance, $\ulogtrace_1$ has two events with uncertainty: $\eid{1}$ and $\eid{2}$. The former is uncertain (confidence $0.25$), has event label $\m a$ (with confidence $1$), timestamp interval $[0,5]$ and a variable assignment such that $x$ is assigned to either $2$ or $3$. Also $\eid{2}$ is uncertain, has label $\m b$ or $\m c$ (with associated confidence values $0.8$ and $0.2$, respectively), timestamp $2$ and variable assignment $y=1$. Another example of an uncertain event is $\eid{3}$ in $\ulogtrace_2$, where $x$ takes a value from the interval $[1,6.5]$.  
\end{example}

An activity label $b\,{\in}\,A$ is \emph{admissible} for an event with uncertainty $\uevent$ iff it is consistent with 
$\ueventLabels(\uevent)$, i.e., 
if there is some $p$ such that $(b,p) \in \ueventLabels(\uevent)$.
Admissibility of timestamp and variable values is defined similarly.

Intuitively, given a log trace with uncertainty $\ulogtrace$, a \emph{realization} of $\ulogtrace$ is a sequence $\logtrace=\seq[n]e$ of events corresponding to a possible sequentialization of \emph{a subset of} the events with uncertainty in $\ulogtrace$ that is consistent with their uncertain timestamps, and in which only one possible value is chosen for event labels and variable assignments. 
%
%
The remaining events with uncertainty in $\ulogtrace$ but not in $\logtrace$ are simply discarded.

An event without uncertainty, or simply \emph{event}, is a tuple $(\ueventId,b, \hat{\alpha})$, where 
$\ueventId$ is again an event identifier, $b\in A$ is an activity label, 
and $\hat{\alpha}$ is a special  variable assignment that assigns to each variable $v\in V$ a \emph{single} value of the correct type.
Given an event $e=(\ueventId,b, \hat{\alpha})$, we denote its components by $\ueventId(e)$, $\lab(e)$ and $\hat{\alpha}(e)$, respectively. 
These events are akin to the standard notion of events in conformance checking literature, extended with variable assignments as in \cite{cocomot}, with the addition of identifiers (which are needed to relate them to the corresponding event with uncertainty in the log, as explained later). 
The set of all possible such events is denoted by $\E$.

\begin{definition}[Realization]
\label{def:realization}
A sequence $\logtrace=\seq[n]e$ of events as above is a \emph{realization} of a log trace with uncertainty $\ulogtrace$ if there is a subset $\sset[n]\uevent \subseteq \ulogtrace$
 and a sequence of timestamps $t_1 \leq t_2 \leq \dots \leq t_n$ such that for each $i\in[1,n]$:
\begin{compactenum}[(i)]
\item $t_i$ is admissible for $\uevent_i$, hence defining an ordering on $\logtrace$; 
\item $\ueventId(e_i)=\ueventId(\uevent_i)$;
\item $\lab(e_i)=b$ with $b$ admissible for $\uevent_i$;
\item 
$\hat{\alpha}(e_i)(v) \in \alpha(\uevent_i)(v)$ for all $v$ such that $\alpha(\uevent_i)(v)$ is defined.
\end{compactenum}
Moreover, we impose that for every $\uevent\in\ulogtrace$ with $\ueventConfidence(\uevent)=1$ there is an event $e\in\logtrace$ with $\ueventId(e_i)=\ueventId(\uevent_i)$, namely a realization cannot discard events in the log that are not uncertain. 
\label{def:prel}	
\end{definition}

A realization of a trace with uncertainty $\ulogtrace$ is thus a possible sequentialization of (a subset of) the events with uncertainty in $\ulogtrace$ in which a single, admissible timestamp value, activity label and value for variables are selected from the corresponding event with uncertainty $\uevent\in\ulogtrace$ with $\ueventId(e)=\ueventId(\uevent)$. 
We denote that $\logtrace$ is a realization of $\ulogtrace$ by writing $\logtrace \in \R(\ulogtrace)$. 
Events in a realization $\logtrace$ are no longer associated with confidence values (which remain in $\ulogtrace$). 

Note that $\R(\ulogtrace)$ cannot be empty, as it is always possible to select $\set{t_1,\ldots,t_n}$ as in Def.~\ref{def:prel}: even if two events cannot be ordered because they admit the same single timestamp, both orderings are accounted for by different realizations. 
$\R(\ulogtrace)$ can be infinite if data variables are assigned by $\ulogtrace$ to intervals over dense domains.

\begin{example}
\label{exa:3}
Consider the trace with uncertainty $\ulogtrace_1$ in Ex.~\ref{exa:2}. It has 13 realizations, since the first event has two possible variable assignments, the second event has two possible labels; moreover, the two events can be ordered in both ways and in addition each event can also be removed (as they are uncertain).

Two possible realizations of $\ulogtrace_1$ are 
$\logtrace' = \tup{\tup{\eid{1},\m a, \{x \mapsto 2\}}, \tup{\eid{2},\m b, \{y \mapsto 1\}}}$ and 
$\logtrace'' = \tup{\tup{\eid{2},\m c, \{y \mapsto 1\}}, \tup{\eid{1},\m a, \{x \mapsto 3\}}}$. 
Note that these realizations differ in the order of the two events, label selection and variable assignments. 
\end{example}

We focus on a conformance checking procedure to construct an \emph{alignment} of a log trace $\logtrace$ (that is a realization of a log trace with uncertainty $\ulogtrace$) w.r.t. the process model (i.e., the DPN $\NN$), by matching event labels in the log trace against transition firings in the process runs of $\NN$. 
However, when constructing an alignment, not every event in the log trace can always be put in correspondence with a transition firing, and vice versa. 
Therefore, as customary, we consider a special ``skip" symbol $\empty$ and the extended set of events $\eventsplus = \E\cup \set{\empty}$ and, given $\NN$, the extended set of transition firings $\firingsplus=\firingsof{\NN}\cup\set{\empty}$. 

Given a DPN $\NN$ and a set  $\E$ of events (without uncertainty) as above, a pair $(e,f)\in \eventsplus \times \firingsplus \setminus \{(\empty,\empty)\}$ is called \emph{move}.
A move $(e,f)$ is called: 
\begin{inparaenum}[(i)]
\item \emph{log move} if $e \in \E$ and $f = \empty$; 
\item \emph{model move} if $e = \empty$ and $f \in \firingsof{\NN}$; 
\item \emph{synchronous move} if $(e,f) \in \E \times \firingsof{\NN}$.
\end{inparaenum}
Let $\moves$ be the set of all such moves. We now show how moves can be used to define alignments of realizations.

For a sequence of moves 
$\gamma = \tup{(e_1,f_1), \dots, (e_n,f_n)}$,
the \emph{log projection} $\restr{\gamma}{L}$ of $\gamma$ is the subsequence $\seq[i]{e'}$ of $\seq[n]e$ that is in $\E^*$ and is obtained by projecting away from $\gamma$ all $\empty$ symbols. Similarly, the \emph{model projection} $\restr{\gamma}{M}$ of $\gamma$ is the subsequence $\seq[j]{f'}$ of $\seq[n]f$ such that $\seq[j]{f'}\in \firingsof{\NN}^*$.

\begin{definition}[Alignment]
Given $\NN$, a sequence of moves $\gamma$
is a \emph{complete alignment} of a realization $\logtrace$ if $\restr{\gamma}{L} = \logtrace$ and $\restr{\gamma}{M}\in\runsof{\NN}$.
\end{definition}

\begin{example}\label{exa:4}
Consider the realization $\logtrace' = \tup{\tup{\eid{1}, \m a, \{x \mapsto 2\}}, \tup{\eid{2}, \m b, \{y \mapsto 1\}}}$ from Ex.\ref{exa:3}. The following are examples of possible complete alignments of $\logtrace'$ with respect to the DPN from Ex.~\ref{exa:1}: 
\begin{align*}
\gamma^1_{\logtrace'}\;&
\syncmove{\eid{1}}{}{\m a}{x^w\mapsto 2}
\syncmove{\eid{2}}{}{\m b}{y^w\mapsto 1}
\modelmove{\m c}{}
&
\gamma^2_{\logtrace'}\;&
\syncmove{\eid{1}}{}{\m a}{x^w\mapsto 5}
\syncmove{\eid{2}}{}{\m b}{y^w\mapsto 1}
\modelmove{\m c}{}
&
\gamma^3_{\logtrace'}\;&
\syncmove{\eid{1}}{}{\m a}{x^w\mapsto 2}
\modelmove{\m b}{y^w\mapsto 2}
\syncmove{\eid{2}}{}{\m d}{}
\end{align*}
\end{example}

\noindent
We denote by $\alignment{\logtrace'}$ the set of all complete alignments for $\logtrace'$ w.r.t. $\NN$.

As shown in Ex.~\ref{exa:4}, some alignments are more fitting than others: for instance, they can have mismatching variable assignments (e.g., in the first move of $\gamma^2_{\logtrace'}$) and label matching (e.g., in the third move of $\gamma^3_{\logtrace'}$).
This will be captured by the cost function, described next.

\section{Costs and Optimal Alignments}
\label{sec:costmodel}

In this paper we do not wish to restrict to specific cost functions, and therefore fix only a \emph{cost schema} which leaves several elements arbitrary. We however illustrate the the cost 
components and describe one possible instantiation of said schema, which we use in the encoding in Sec.~\ref{sec:encoding}. 
The overall cost schema for alignments is shown in Fig.~\ref{fig:costs}. 

We first give the intuition. 
The general idea is that, as we are not merely interested in finding a cost-minimal alignment for
an arbitrary
realization as in~\cite{PegoraroUA21}, i.e., without considering the confidence associated to the selection of realizations, we impose a confidence cost on realizations \emph{in addition} to the cost of aligning them, as illustrated in Fig.~\ref{fig:costs}. 
As a result, the cost $\mathfrak{K}(\gamma_\logtrace,\ulogtrace)$ of an alignment
$\gamma_\logtrace$ with respect to an uncertain trace $\ulogtrace$
is the sum of two costs:

\begin{figure}[t]
\centering
\begin{tikzpicture}[]
\tikzstyle{node}=[yshift=4mm, scale=1.1, transform shape]
\node[node] at (0,0) {
$\mathfrak{K}(\gamma_\logtrace,\ulogtrace) 
= 
\overbrace{\sum_{i\in[1,n]} \underbrace{\kappa(e_i,f_i)}_{\substack{\text{data-aware} \\ \text{alignment cost~(Sec.~\ref{ssec:distance}) }}} \underbrace{\theta(e_i,\ulogtrace)}_{\substack{\text{\; confidence cost} }}}^{\text{alignment cost } \kappa_A(\gamma_\logtrace,\ulogtrace)}$
+
$\overbrace{\sum_{e\in \ulogtrace, e\not\in\logtrace} \kappa_{\ulogtrace}(e)}^{\text{event removal cost } \kappa_R(\logtrace,\ulogtrace)}$
};
\node[node] at (-0.06,0.05) {$\otimes$};
\end{tikzpicture}
\caption{Structure of the cost of an alignment $\gamma_\logtrace=\tup{(e_1,f_1), \dots, (e_n,f_n)}$ of a realization $\logtrace$ of a trace with uncertainty $\ulogtrace$. The cost associated to the selection of $\logtrace$ is given by $\kappa_R(\logtrace,\ulogtrace)$ plus, at each step, the additional penalty given by $\theta(e_i,\ulogtrace)$ according to $\otimes$.}
\label{fig:costs}	
\end{figure}

\smallskip
\noindent
\textbf{1) The alignment cost} $\kappa_A(\gamma_\logtrace,\ulogtrace)$ measures the quality of the alignment $\gamma_\logtrace$ for the realization $\logtrace$. As customary in the conformance checking literature, it is 
based on a mapping $\kappa\colon \moves \to \mathbb R^+$ that assigns a cost to every move $(e_i,f_i)\in\gamma_\logtrace$. 
In Sec.~\ref{ssec:distance} we will discuss in more detail how this function $\kappa$ can be defined. 

In addition, for synchronous moves and log moves, this cost is combined with a confidence penalty that depends on $\ueventConfidence(e_i)$ and on the confidence value $p$ associated to the activity label $b=\lab(e_i)$ according to the event with uncertainty $\uevent$ so that $\ueventId(e_i)=\ueventId(\uevent)$, i.e., $(b,p)\in \ueventLabels(\uevent)$. 
Intuitively, this imposes a penalty for selecting $b$ as the activity
chosen for 
$e_i$ in the realization $\logtrace$ of $\ulogtrace$.

We do not fix a specific calculation of this penalty, but keep it parametric and denote it as $\theta(e_i,\ulogtrace)$. 
The cost of an alignment $\gamma_\logtrace$ can then be defined as: 
\begin{equation}
\label{eq:costsum}
\kappa_A(\gamma_\logtrace,\ulogtrace) = \textstyle\sum_{i=1}^n \kappa(e_i,f_i)\otimes \theta(e_i,\ulogtrace)
\end{equation}
\noindent
where $\otimes$ denotes an arbitrary operator to combine the two costs. 
%


For instance, in Sec.~\ref{sec:encoding} we assume, for a realization $\logtrace$ of $\ulogtrace$ and alignment $\gamma_\logtrace=\tup{(e_1,f_1), \dots, (e_n,f_n)}$:
\begin{align}
\label{eq:costtheta}
\kappa(e_i,f_i)\otimes \theta(e_i,\ulogtrace)&=
\left\{
\begin{array}{ll}
\kappa(e_i,f_i)  & \text{if } e_i=\empty, \text{ otherwise:}\\
\theta(e_i,\ulogtrace)  & \text{if } \kappa(e_i,f_i)=0 \\
\kappa(e_i,f_i)\cdot (1+\theta(e_i,\ulogtrace)) \quad & \text{if } \kappa(e_i,f_i)>0 \\      
\end{array}\right.
\end{align}
in which we fix $\theta(e_i,\ulogtrace)=(1-\ueventConfidence(e_i))+(1-p)$, where $b$ is the label of $e_i$, i.e., $b=\lab(e_i)$, and $p$ is the confidence value associated to $b$, i.e., $(b,p)\in \ueventLabels(\uevent)$.%

Intuitively, in this definition of $\kappa_A(\gamma_\logtrace,\ulogtrace)$, the cost of model moves is simply (a data-aware extension of) the usual alignment cost,  which we define in Sec.~\ref{ssec:distance}. Otherwise, the cost includes a penalty for having selected $\lab(e_i)$ in the realization $\logtrace$ of $\ulogtrace$. Such penalty decreases the more we are confident about the selected activity among the possible activities associated to the event with uncertainty. 
Other definitions of $\theta$ and $\otimes$ are however possible.

\smallskip
\noindent
\textbf{2) The event removal cost} $\kappa_R(\logtrace,\ulogtrace)$ measures the cost of selecting the subsets of the events in $\ulogtrace$
that appear in $\logtrace$, discarding the remaining (uncertain) events. 
Although we do not wish to restrict to a specific function $\kappa_R$, a reasonable option is to assume it to be based on a mapping $\kappa_{\ulogtrace}\colon \E \to \mathbb R_{\geq 0}$ that assigns a removal cost to each event, proportionally to the confidence value $\ueventConfidence(\uevent)$ for $\uevent\in\ulogtrace$ so that $\ueventId(e)=\ueventId(\uevent)$. 
Hence, the total event removal cost can be computed as:  
\[
\kappa_R(\logtrace,\ulogtrace)= \textstyle\sum_{e\in \ulogtrace, e\not\in\logtrace}~ \kappa_{\ulogtrace}(e)
\]
For instance, in Sec.~\ref{sec:encoding} we will take $\kappa_{\ulogtrace}(e)$ to be precisely  $\ueventConfidence(\uevent)$, for $\uevent$ as above, when such a confidence value is less than $1$, and equal to infinity otherwise (to prevent events that are not indeterminate to be discarded from realizations).  
Other definitions of $\kappa_R$ are however possible.
Again, according to these expressions, the cost of selecting $\logtrace$ as a realization of $\ulogtrace$ results from $\kappa_R(\logtrace,\ulogtrace)$ for removed events plus, at each step, a penalty $\theta(e_i,\ulogtrace)$ for not having discarded  $e_i$ but having selected one admissible label among those associated to the uncertain event in $\ulogtrace$ with the same $\ueventId$. 

\begin{example}
\label{exa:5}
Consider again the trace with uncertainty $\ulogtrace_1$ from Example~\ref{exa:3}:
  
\begin{small}
$
\begin{array}{c}
\ulogtrace_1 = \{
 \tup{\eid{1},.25, \set{\m a\,{:}\,1}, [0\text{-}5], \{x \mapsto \set{2,3}\}},\;
 \tup{\eid{2},.9, \set{\m b\,{:}\,.8, \m c\,{:}\,.2}, \{2\}, \{y \mapsto \set{1}\}}\} 
 \end{array}
$
\end{small}

\noindent
and three of its possible realizations $\logtrace_1 = \tup{\tup{\eid{1}, \m a, \{x \mapsto 3\}}}$, 
 $\logtrace_2 = \tup{\tup{\eid{2},\m b, \{y \mapsto 1\}}}$ and $\logtrace_3 = \tup{\tup{\eid{2},\m c, \{y \mapsto 1\}}}$, where in all cases one of the two events was removed. 
If we adopt the specific implementation of cost functions exemplified above (and used in our encoding in Section~\ref{sec:encoding}), we have that $\kappa_R(\logtrace_2,\ulogtrace_1)>\kappa_R(\logtrace_1,\ulogtrace_1)$ since $\ueventConfidence(\eid{2})>\ueventConfidence(\eid{1})$. Similarly, the difference between $\logtrace_2$ and $\logtrace_3$ is only in the activity chosen for $\eid{2}$, therefore the cost of selecting $\logtrace_2$ is smaller than that for $\logtrace_3$, because the confidence associated to activity $\m b$ is greater than the one associated to $\m c$; hence  $\theta(\tup{\eid{2},\m b, \{y \mapsto 1\}},\ulogtrace_1)<\theta(\tup{\eid{2},\m c, \{y \mapsto 1\}},\ulogtrace_1)$. 
 \end{example} 
 
\begin{definition}[Cost of alignments]
Fixed the two arbitrary cost functions $\kappa_A$ and $\kappa_R$ introduced above, 
given $\NN$, a trace with uncertainty $\ulogtrace$ that has realization $\logtrace=\seq[n]{e}$ and an alignment $\gamma_\logtrace = \tup{(e_1,f_1), \dots, (e_n,f_n)} \in \alignment{\logtrace}$, the \emph{cost} of $\gamma_\logtrace$ w.r.t. $\ulogtrace$, denoted $\mathfrak{K}(\gamma_\logtrace,\ulogtrace)$, is obtained as shown in Figure~\ref{fig:costs}:
\begin{center}
$\mathfrak{K}(\gamma_\logtrace,\ulogtrace) = \kappa_A(\gamma_\logtrace,\ulogtrace) + \kappa_R(\logtrace,\ulogtrace).$
\end{center}
\label{def:tot_cost}
\end{definition}

An alignment $\gamma_\logtrace$ is \emph{optimal for $\logtrace$} if $\kappa_A(\gamma_\logtrace,\ulogtrace)$ is minimal among all complete alignments for $\logtrace$, i.e., there is no $\gamma'_\logtrace\in \alignment{\logtrace}$ with $\kappa_A(\gamma'_\logtrace,\ulogtrace)<\kappa_A(\gamma_\logtrace,\ulogtrace)$. 
Similarly, given $\NN$ and a trace with uncertainty $\ulogtrace$, we say that $\gamma_\logtrace$ is \emph{optimal for} $\ulogtrace$ if $\mathfrak{K}(\gamma_\logtrace,\ulogtrace)$ is minimal among all possible realizations of $\ulogtrace$, i.e., there is no other realization $\logtrace'\in\R(\ulogtrace)$ and alignment $\gamma_{\logtrace'}\in \alignment{\logtrace'}$ so that $\mathfrak{K}(\gamma_{\logtrace'},\ulogtrace) < \mathfrak{K}(\gamma_\logtrace,\ulogtrace)$. 

\begin{definition}[Conformance checking]
\label{def:conformance_checking}
Given $\NN$, the \emph{conformance checking task} for a trace with uncertainty $\ulogtrace$ is to find a realization $\logtrace$  of $\ulogtrace$ and an alignment $\gamma_\logtrace$ that is optimal for $\ulogtrace$. 
\end{definition}

\noindent 
Multiple realizations $\logtrace$ and optimal alignments $\gamma_\logtrace$ may exist
for $\ulogtrace$, though the minimal cost is unique for a given cost function. 
The \emph{conformance checking task for an unordered log} consists of the
conformance checking task for all its traces.

Note that we can easily formulate the task of finding the lower-bound on the cost of possible alignments among all realizations (as in~\cite{PegoraroUA21}), given $\ulogtrace$, by simply imposing $\kappa_R(\logtrace,\ulogtrace)=0$, $\theta(e,\ulogtrace)=1$ and by taking $\otimes$ as product: this corresponds to impose no cost for selecting an arbitrary realization, thus simply returning one that has minimal alignment cost $\kappa$.  

In the remainder, we discuss separately the definition of alignment cost $\kappa$.

\subsection{Data-aware Alignment Cost Function}
\label{ssec:distance}

We use a generalized form of a cost function to measure the conformance between a realization and a process run in $\runsof{\NN}$, i.e., to define $\kappa\colon \moves \to \mathbb R_{\geq 0}$ used in Def.~\ref{def:tot_cost}. 
As in~\cite{cocomot}, we parameterize this by three penalty functions:
\[
P_L\colon \E \to \mathbb N \qquad
P_M\colon \firingsof{\NN} \to \mathbb N \qquad 
P_=\colon \E \times \firingsof{\NN} \to \mathbb N \\
\]
\noindent
called \emph{log move penalty}, \emph{model move penalty} and \emph{synchronous move penalty}, respectively.
Intuitively, $P_L(e)$ gives the cost that has to be paid for a log move $e$;
$P_M(f)$ penalizes a model move $f$; and
$P_=(e,f)$ expresses the cost to be paid for a synchronous move of $e$ and $f$. 
By suitably instantiating $P_=$, $P_L$, and $P_M$, one can obtain conventional cost functions~\cite{cocomot}:
the Levenshtein distance~\cite{BoltenhagenCC19,BoltenhagenCC21}, standard cost function for multi-perspective conformance checking~\cite{MannhardtLRA16,Mannhardt18}.  

Then, the data-aware cost function $\kappa\colon \moves \to \mathbb R_{\geq 0}$ we adopt in Def.~\ref{def:tot_cost} is simply defined as $\kappa(e,f)=P_L(e)$ if $f=\empty$, $\kappa(e,f)=P_M(f)$ if $e=\empty$, and $\kappa(e,f)=P_=(e,f)$ otherwise. 

\paragraph{Data-aware Cost Component of $P_=$.} 
Crucially, for DPNs we typically
consider a data-aware extension of the usual distance-based cost function for synchronous moves. 
Indeed, given an event $e=(\ueventId,b, \hat{\alpha})$ of a realization and a transition firing $f=(t,\beta)$, we want $P_=(e,f)$ to compare also the values assigned to variables by $\hat{\alpha}$ and $\beta$.
For instance, in Ex.~\ref{exa:4}, the alignment $\gamma^2_{\logtrace_1}$ is so that its first (synchronous) move has a mismatch between the value assigned to variable $x$ by the event $\eid{1}$ (i.e., $\hat{\alpha}(\eid{1})(x)=2$) and transition firing $(a,\set{x^w\mapsto 5})$.
Various data-aware realizations of $P_=$ have been already addressed in the literature~\cite{MannhardtLRA16,cocomot}.

\begin{example}
\label{exa:6}
Consider again the trace with uncertainty $\ulogtrace_1$ from Ex.~\ref{exa:5}, i.e., 
$\small
\ulogtrace_1 = \{
 \tup{\eid{1},.25, \set{\m a\,{:}\,1}, [0\text{-}5], \{x \mapsto \set{2,3}\}},\;
 \tup{\eid{2},.9, \set{\m b\,{:}\,.8, \m c\,{:}\,.2}, \{2\}, \{y \mapsto \set{1}\}}\} 
$. 
Assume to fix $P_M$, $P_L$ to be as usual in 
the standard cost function, as illustrated in~\cite{cocomot}, namely 
$P_L(b,\alpha) = 1$;
$P_M(t,\beta) = 0$ if $t$ is silent (i.e., $\ell(t) = \tau$) and 
$P_M(t,\beta)$ equal to 1 plus the number of variables written by $guard(t)$ otherwise. 
For $P_=$, assume a data-aware extension (of the $P_=$ used to match the standard cost function~\cite{cocomot}) defined as: 
$P_=(\tup{\ueventId,b, \hat{\alpha}}, (t,\beta)) = |\{v \mid \hat{\alpha}(v) \neq \beta(v^w)\}| \:/\: |V| $ if $b$ is the label of $t$, i.e. $b = \ell(t)$, and
$P_=(\tup{\ueventId,b, \hat{\alpha}}, (t,\beta)) = \infty$ otherwise. 
Then, if we instantiate cost functions as in Ex.~\ref{exa:5} (also used in our encoding in Sec.~\ref{sec:encoding}), the optimal alignment of $\ulogtrace_1$ w.r.t. the DPN $\NN$ depicted in Ex.~\ref{exa:1} is $\gamma^1_{\logtrace'}$ as shown in Ex.~\ref{exa:4} (of cost 2.05). 

Further, if we consider the task of finding the lower-bound on the cost of optimal alignments for any realization of $\ulogtrace_1$ (as discussed below Def.~\ref{def:conformance_checking}), then this is 1 and it is given 
as well by the realization $\logtrace'$ and $\gamma^1_{\logtrace'}$.
\end{example}


\section{Encoding}
\label{sec:encoding}

In this section we describe our SMT encoding, obtained as the result of 4 steps:
\begin{compactenum}[(1)]
\item represent the process run, the trace realization, and the alignment symbolically by a set of SMT variables;
\item set up constraints $\Phi$ that express optimality of the alignment;
\item solve $\Phi$ to obtain a satisfying assignment $\nu$; 
\item decode the process run, trace realization, and optimal alignment $\gamma$ from $\nu$.
\end{compactenum}
The same procedure was followed in \cite{cocomot}, with important differences.  
In step (1), we now need to represent both the process run and also the trace realization, which is
complicated by the fact
that the order of the events is not fixed. Moreover, the cost functions are defined differently, as described in Sec.~\ref{sec:costmodel}. 
These changes also affect the decoding in step (4).

Similarly to earlier SAT-based approaches~\cite{BoltenhagenCC21,cocomot}, we aim to construct a symbolic representation of both a process run and an alignment, that are subsequently concretized using an SMT solver. 
Since the symbolic representation depends on a finite set of initial variable declarations (and thus must be finite), we need to fix upfront an upper bound on the size of the process run. This upper bound, and even its existence, depends on the cost function of choice.
The  Lemma below shows how a (coarse) upper bound can be established for the cost model from Sec.~\ref{sec:costmodel},
where the cost function is the standard one as in Ex.~\ref{exa:6}.

\begin{lemma}
\label{lem:bound}
Let $\NN$ be a DPN and $\ulogtrace$ a trace with uncertainty 
that has $m_1$ certain and $m_2$ uncertain events.
Let $\seq[n]f$ be a run of $\NN$ such that 
$c = \sum_{j=1}^n P_M(f_j)$ is minimal,
and $k$ the length of the longest acyclic sequence of silent transitions in $\NN$.
Then there is an optimal alignment $\gamma$ for $\ulogtrace$ such that
the length of $\restr{\gamma}{M}$ is at most $(4m_1 + 2m_2 + c)\cdot k$.
\end{lemma}
\longversion{
\begin{proof}
Let $\gamma_0=\langle (e_1, \empty), \dots, (e_{m}, \empty), (\empty, f_1), \dots (\empty, f_n)\rangle$ be a valid alignment for $\tup{e_1, \dots, e_{m}}\in\R(\ulogtrace)$, where $m=m_1+m_2$.
Its cost is computed as follows. 
First, for the log steps $\gamma_0' = \langle(e_1, \empty), \dots, (e_{m}, \empty)\rangle$
we have $\kappa_A(\gamma_0',\ulogtrace) + \kappa_R(\gamma_0',\ulogtrace)$.
Since for each uncertain event $e$ in $\ulogtrace$ the event removal cost is $\kappa_{\ulogtrace}(e)< 1$, then $\kappa_R(\gamma_0',\ulogtrace) \leq m_2$.
Second, $\kappa_A(\gamma_0',\ulogtrace) = \textstyle\sum_{i=1}^{m_1} \bigl(\kappa(e_i,\empty)\cdot (1+\theta(e_i,\ulogtrace))\bigr)$,
where $\kappa(e_i,\empty) = 1$ and $\theta(e_i,\ulogtrace) \leq 2$. 
Hence, $\kappa_A(\gamma_0',\ulogtrace) \leq 3m_1$.
Then, overall, $\mathfrak{K}(\gamma_0,\ulogtrace) \leq 3m_1 + m_2 + c$, where $c$ is the cost of the model steps (by assumption).

To be optimal, $\gamma$ must satisfy 
$\mathfrak{K}(\gamma,\ulogtrace) \leq \mathfrak{K}(\gamma_0,\ulogtrace)$.
By assumption, $\gamma$ has at most $m$ synchronous moves. 
For simplicity and a conservative estimate, we assume their cost is $0$.
In addition, $\restr{\gamma}{M}$ may feature non-silent moves, each costing at least $1$, 
and thus have at most $3m_1 + m_2 + c$ non-silent moves (otherwise, we would have $\mathfrak{K}(\gamma,\ulogtrace) > \mathfrak{K}(\gamma_0,\ulogtrace)$). 
Thus $\gamma$ has at most $4m_1 + 2m_2 + c$ synchronous moves and model moves corresponding to non-silent transitions.
However, in between every one of these, as well as before and afterwards,
there may be silent transitions that have by assumption cost 0.
There could also be loops which consist of silent transitions only, and executing such a loop an arbitrary number of times does not incur any additional cost.
However, as silent transitions do not write variables, an alignment whose process run involves such a loop cannot have strictly smaller cost than  the alignment obtained by omitting the loop.
So by assumption, it is safe to assume that in the optimal alignment in between two non-silent transitions there are at most $k$ silent ones.
Thus, the length of $\restr{\gamma}M$ is at most $(4m_1 + 2m_2 + c)\cdot k$.
\qed
\end{proof}}
{The proof of this lemma can be found in~\cite{BPMArxiv22}.}
Note that, in case the model admits loops that entirely consist of silent transitions, then there can be infinitely many optimal alignments that are not bounded in length (as such loops can be repeated arbitrarily many times without incurring in any additional penalty on the alignment cost). Thus, the above lemma shows only \emph{existence} of an optimal alignment within that bound, but in general the bound does not apply to \emph{all} optimal alignments.

\subsection{Encoding the Process Run}

Assuming that the process run in the optimal alignment has length at most $n$, 
we use the following SMT variables to represent this run:
\begin{compactitem}
\item[(a)]
transition step variables $\transvar_i$ for $1\leq i \leq n$
of type integer; if $T = \{t_1, \dots, t_{|T|}\}$ then it is ensured
that $1\,{\leq}\,\transvar_i\,{\leq}\,|T|$, so that
$\transvar_i$ is assigned $j$ iff the $i$-th
transition in the process run is $t_j$;
\item[(b)]
marking variables $\markvar_{i,p}$ of type integer for all $i$, $p$ with
$0\leq i \leq n$ and $p\in P$, where $\markvar_{i,p}$ is assigned $k$ iff
there are $k$ tokens in place $p$ at instant $i$;
\item[(c)]
data variables $\datavar_{i,v}$ for all $v\in V$ and $i$, $0\leq i \leq n$;
the type of these variables depends on $v$, 
with the semantics that $\datavar_{i,v}$ is assigned $r$ iff the value of $v$ at 
instant $i$ is $r$;
we also write $\datavar_{i}$ for $(\datavar_{i,v_1}, \dots, \datavar_{i,v_k})$.
\end{compactitem}
\noindent
Note that variables (a)--(c) encode all information required to capture a process run of a DPN with $n$ steps. They will be used to represent the model projection
of the alignment $\gamma$. 
To encode the process run, we use the constraints 
\[\varphi_{\mathit{run}} = \varphi_{\mathit{init,fin}} \wedge
\varphi_{\mathit{trans}} \wedge
\varphi_{\mathit{enabled}} \wedge
\varphi_{\mathit{mark}} \wedge
\varphi_{\mathit{data}}\]
where the subformulas above reflect requirements to the solution as follows:
\begin{compactitem}
\item 
The initial and final markings $M_I$ and $M_F$, and the initial assignment $\alpha_0$ are respected:
\begin{align}
\tag{$\varphi_{\mathit{init,fin}}$}
&\textstyle\bigwedge_{p\in P} \markvar_{0,p}\,{=}\, M_I(p) \wedge
\textstyle\bigwedge_{v\in V} \datavar_{0,v} \,{=}\,  \alpha_0(v)  \wedge 
\textstyle\bigwedge_{p\in P} \markvar_{n,p} \,{=}\,  M_F(p)  &
\end{align}
\item
Transitions correspond to transition firings in the DPN:
\begin{align}
\label{eq:transition range}
\tag{$\varphi_{\mathit{trans}}$}
\textstyle\bigwedge_{1\leq i \leq n} 1 \leq \transvar_i \leq |T|
\end{align}
\item Transitions are enabled when they fire:
\begin{align}
\label{eq:enabled}
\tag{$\varphi_{\mathit{enabled}}$}
\textstyle\bigwedge_{1\leq i \leq n} \bigwedge_{1\leq j \leq |T|}
{(\transvar_i\,{=}\,j)} \to 
\textstyle\bigwedge_{p\,\in\,\pre{t_j}} \markvar_{i-1,p} \geq |\pre{t_j}|_p 
\end{align}
where $|\pre{t_j}|_p$ denotes the multiplicity of $p$ in the multiset $\pre{t_j}$.
\item We encode the token game:
\begin{align}
\label{eq:token game}
\tag{$\varphi_{\mathit{mark}}$}
\bigwedge_{1\leq i \leq n} \bigwedge_{1\leq j \leq |T|}
{(\transvar_i\,{=}\,j)} \to 
\bigwedge_{p\,\in\,P} \markvar_{i,p} - \markvar_{i-1,p} =  |\post{t_j}|_p -  |\pre{t_j}|_p 
\end{align}
where $|\post{t_j}|_p$ is the multiplicity of $p$ in the multiset $\post{t_j}$.
\item The transitions satisfy the constraints on data:
\begin{align}
\label{eq:data}
\tag{$\varphi_{\mathit{data}}$}
\bigwedge_{1\leq i < n} \bigwedge_{1\leq j \leq |T|}
{(\transvar_i\,{=}\,j)} \to 
guard(t_j)\chi \wedge
\bigwedge_{v\not \in write(t_j)} \datavar_{i-1,v} = \datavar_{i,v}
\end{align}
where the substitution $\chi$ uniformly replaces $V^r$ by $\datavar_{i-1}$ and $V^w$ by $\datavar_i$. Above, $write(t)$ denotes the set of variables that are written by $guard(t)$. 
\end{compactitem}

\subsection{Trace Realization Constraints}
\label{ssec:realization}

Next, we describe how an admissible realization for a given trace with uncertainty $\ulogtrace$ is encoded.
To this end, additional variables are needed.
Let $\ulogtrace = \{\uevent_1, \dots, \uevent_m\}$ such that 
$\uevent_i = \tup{\ueventId,\ueventConfidence,\ueventLabels,\ueventTss,\alpha}$ for each $1 \leq i \leq m$,  with 
$\ueventLabels = \set{b_1:p_1, \ldots, b_{N_i}:p_{N_i}}$. 
We use the following sets of variables for all $i$:
\begin{compactitem}
\item[(d)] a boolean \emph{drop variable} $\dropvar_{\uevent_i}$ expressing whether the event is absent in the realization; it must satisfy $\dropvar_{\uevent_i} \Longrightarrow (\comp{\uevent_i}{\ueventConfidence}\,{<}\,1)$,
i.e., it can only be assigned true for uncertain events with confidence below 1, 
\item[(e)] an integer \emph{activity variable} $\actvar_{\uevent_i}$ that expresses which of the labels $b_1, \dots, b_{N_i}$ is taken, so it must satisfy
$1 \leq \actvar_{\uevent_i} \leq N_i$, and
\item[(f)] \emph{trace data variables} $\tdvar_{v,\uevent_i}$ of suitable type for all $v\in V$ 
that satisfy either that
$\bigvee_{c \in \uevent.\alpha}\tdvar_{v,\uevent_i}=c$ if $\alpha(\uevent)$ is a set, or
$l\,{\leq}\,\tdvar_{e_i}\,{\leq}\,u$ if $\alpha(\uevent) = [l,u]$ is an interval.
\end{compactitem}

If each uncertain event  in $\ulogtrace$ has a single, distinct timestamp,
we call $\ulogtrace$ \emph{sequential}, and assume it is ordered by time as
$\seq[n]\uevent$.
If $\ulogtrace$ is not sequential, we need the following additional variables:
For all $i$, $1 \leq i \leq m$: 
\begin{compactitem}
\item[(g)] a \emph{time stamp variable} $\tsvar_{\uevent_i}$ to express when event $\uevent_i$ happened, with the constraint
$\bigvee_{t\in \ueventTss}\tsvar_{\uevent_i}=t$ if $\ueventTss(\uevent_i)$ is a set, or
$l\,{\leq}\,\tsvar_{e_i}\,{\leq}\,u$ if $\ueventTss(\uevent_i)= [l,u]$ is an interval, 
\item[(h)] an integer \emph{position variable} $\posvar_{\uevent_i}$ to fix the position of $\uevent_i$ in the realization,
\item[(i)] an integer \emph{item variable} $\nthvar_j$ that indicates the $j$-th element in the realization, i.e., $\nthvar_j$ has value $\ueventId(\uevent_i)$ if and only if the $j$-th event in the trace with uncertainty is $\uevent_i$; we thus issue the constraint $\bigvee_{i=1}^m \nthvar_j = \ueventId(\uevent_i)$ to fix the range of $\nthvar_j$, for all $1\leq j \leq m$.
\end{compactitem}
The formula $\varphi_{\mathit{trace}}$ consists of
the range constraints in (d)-(i), in addition to
\begin{align*}
&\textstyle\bigwedge_{i=1}^m \bigwedge_{j=1}^m (\posvar_{\uevent_i} < \posvar_{\uevent_j} \Longrightarrow \tsvar_{\uevent_i} \leq \tsvar_{\uevent_j}) \wedge
  (\tsvar_{\uevent_i} < \tsvar_{\uevent_j} \Longrightarrow \posvar_{\uevent_i} < \posvar_{\uevent_j})\\
&\textstyle\bigwedge_{i=1}^m \bigwedge_{j=1}^m \nthvar_i = \ueventId(\uevent_j)\Longleftrightarrow \posvar_{\uevent_j} = i
\end{align*}
so as to require that, first, the positions assigned to uncertain events by $\posvar_{\uevent_j}$ is compatible with the time stamps assigned by $\tsvar_{\uevent_j}$ and, second, that the $\posvar_{\uevent_j}$ variables work as an ``inverse function'' of the $\nthvar_i$.

\subsection{Encoding the Cost Function}
To encode the alignment and its cost we use, additionally:
\begin{compactitem}
\item[(j)]
distance variables $\distvar_{i,j}$ of type integer
for $0\leq i \leq m$ and $0\leq j \leq n$, where
$\distvar_{i,j}$ is the alignment cost of
the prefix $\logtrace|_i$ of the log trace realization $\logtrace$ and prefix $\procrun|_j$ of the process run $\procrun$, both of which are yet to be determined.
\end{compactitem}

\noindent
The search for an optimal alignment is based on a notion of \emph{edit distance}, similar as in~\cite{cocomot,BoltenhagenCC21}.
More precisely, we assume that the data-aware alignment cost $\kappa(e_i, f_i)$ in Fig. \ref{fig:costs} can be encoded using a \emph{distance-based} cost function with \emph{penalty functions}
$P_L$, $P_M$, and $P_=$ as discussed in Sec.~\ref{ssec:distance}. 
Recall that $P_=$ is assumed to be data-aware, i.e., to take into account the mismatching variable assignments between the events in realizations and transition firings in process runs. 
Intuitively, such functions assess the degree of ``closeness'' between a process run and a log trace.
We assume that there are SMT encodings of these penalty functions that use variables (a)--(i), denoted as $[P_=]_{i,j}$, $[P_M]_{j}$, and $[P_L]_{i}$. 

Moreover, we assume that there are encodings of the event removal cost function $[\kappa_{\ulogtrace}]_i$
and the confidence cost function $[\theta_{\ulogtrace}]_i$, defined for the $i$-th element of the log trace realization.
We then consider the following constraints for $i,j> 0$:%
\footnote{We assume that $P_L$ is always positive, otherwise, a case distinction using $\ite$
is also required in the second line.}
\begin{equation}
\label{eq:delta}
\begin{array}{rl@{\qquad}rl@{\qquad}rl@{\qquad\quad}r}
\distvar_{0,0} &= 0 &
\distvar_{{i},0} &= \min([P_L]_{i}\cdot [\theta_{\ulogtrace}]_i, [\kappa_{\ulogtrace}]_i) + \distvar_{i-1,0} &
\distvar_{0,{j}} &= [P_M]_{j} + \distvar_{0,j-1} 
&\tag{$\varphi_\delta$}\\[1ex] 
\distvar_{i,j} &\multicolumn{5}{l}{=
\min 
\begin{cases}
ite([P_=]_{i, j} = 0, [\theta_{\ulogtrace}]_i, [P_=]_{i, j} + [P_=]_{i, j}\cdot [\theta_{\ulogtrace}]_i) + \distvar_{i-1,j-1}\\
[P_L]_{i}\cdot [\theta_{\ulogtrace}]_i + \distvar_{i-1,j}\\
[\kappa_{\ulogtrace}]_i + \distvar_{i-1,j}\\
[P_M]_{j} + \distvar_{i,j-1}
\end{cases}
}
\end{array}
\notag
\end{equation}
This encoding constitutes an \emph{operational way} for computing the cost function represented in Fig.~\ref{fig:costs}, where the components $\kappa_{\ulogtrace}$ and $\theta$ are distributed
to single moves, which at the same time allows us to use the encoding schema based on the edit distance. The inductive case $\distvar_{i,j}$ is computed so as to locally choose the move with minimal cost. In particular, the first and the second line of the case distinction  correspond exactly to the specific instantiation of the expression $\kappa(e_i,f_i)\otimes \theta(e_i,\ulogtrace)$ exemplified in Sec.~\ref{sec:costmodel}. For instance, the cost penalty $\kappa(e_i,f_i)\cdot (1+\theta(e_i,\ulogtrace))$ in case $\kappa(e_i,f_i)>0$ (see Sec.~\ref{sec:costmodel}) corresponds here, in the $ite$ construct, to the cost penalty $[P_=]_{i, j} + [P_=]_{i, j}\cdot [\theta_{\ulogtrace}]_i$ in the \emph{else} statement.
The expression $\distvar_{m,n}$ encodes then the cost of the complete alignment, which will thus be used as the minimization objective.

The encodings of the penalties, as well as $[\kappa_{\ulogtrace}]_i$ and $[\theta_{\ulogtrace}]_i$, also depend on the choice of the respective functions.
For those exemplified in Sec.~\ref{sec:costmodel}, one can define $[\kappa_{\ulogtrace}]_i$
as a (nested) case distinction on the element from $\ulogtrace$ that is chosen for the $i$-th position (represented with variable $L_i$ -- see Sec.~\ref{ssec:realization}):
\begin{align}
\label{eq:kappaenc}
[\kappa_{\ulogtrace}]_i = &\ite(L_i = \ueventId(\uevent_1) \wedge \dropvar_{\uevent_1},\ueventConfidence(\uevent_1), \dots \\
&\ite(L_i = \ueventId(\uevent_{m}) \wedge \dropvar_{\uevent_{m}},\ueventConfidence(\uevent_m),\infty)\dots ) \notag
\end{align}
A similar case distinction can be done for $[\theta_{\ulogtrace}]_i$, also exemplified in Sec.~\ref{sec:costmodel}.

\subsection{Solving and Decoding}
\label{ssec:decoding}

We use an SMT solver to obtain a satisfying assignment $\nu$ for the 
following constrained optimization problem:
\begin{align}
\label{eq:constraints}
\varphi_{\mathit{run}} \wedge
\varphi_{\mathit{trace}} \wedge
\varphi_{\delta}
\text{\quad minimizing\quad }\distvar_{m,n}
\tag{$\Phi$}
\end{align}
For a satisfying assignment $\nu$ for \eqref{eq:constraints},
we construct the process run 
$\procrun_\nu =  \tup{f_1, \dots, f_n}$
where $f_i = (t_{\nu(\transvar_i)}, \beta_i)$, assuming that the set of transitions $T$ consists of $t_1, \dots, t_{|T|}$ in the ordering already used for the encoding.
The transition variable assignment $\beta_i$ is obtained as follows:
Let the state variable assignments $\alpha_j$, $0\,{\leq}\,j\,{\leq}\,n$, be given by
$\alpha_j(v) = \nu(\datavar_{j,v})$ for all $v\in V$.
Then, $\beta_i(v^r) = \alpha_{i-1}(v)$ and $\beta_i(v^w) = \alpha_{i}(v)$ for all $v\in V$.
Moreover, we construct a realization $\logtrace_\nu=  \tup{e_1, \dots, e_k}$ by ordering the events in $\ulogtrace$
according to $\nu(\tsvar_{\uevent_i})$, dropping those where $\dropvar_{\uevent_i}$ is true,
and fixing the label and data values to $\nu(\actvar_{\uevent_i})$ and $\nu(\tdvar_{\uevent_i})$,
respectively. Finally,
let the (partial) alignments $\gamma_{i,j}$ be defined as follows, for $i,j > 0$:
\begin{align*}
\gamma_{0,0} &=\epsilon \qquad
\gamma_{0,j+1}= \gamma_{0,j} \cdot (\empty, f_{j+1}) \\
\gamma_{i+1,0} &=
\begin{cases}
\gamma_{i,0} \cdot (e_{i+1}, \empty) &
 \text{ if }\nu(\delta_{i+1,0}) = \nu([P_L]_{i+1}\cdot [\theta_{\ulogtrace}]_{i+1} + \delta_{i,0}) \\
\gamma_{i,0} &
 \text{ if }\nu(\delta_{i+1,0}) = \nu([\kappa_{\ulogtrace}]_{i+1} + \delta_{i,0})
\end{cases}\\
\gamma_{i+1,j+1} &= 
\begin{cases}
\gamma_{i,j+1} \cdot (e_{i+1}, \empty) &
 \text{ if }\nu(\delta_{i+1,j+1}) = \nu([P_L]_{i+1}\cdot [\theta_{\ulogtrace}]_{i+1} + \delta_{i,j+1}) \\
\gamma_{i,j+1} &
 \text{ if }\nu(\delta_{i+1,j+1}) = \nu([\kappa_{\ulogtrace}]_{i+1} + \delta_{i,j+1}) \\
\gamma_{i+1,j} \cdot (\empty, f_{j+1}) &
 \text{ if otherwise }\nu(\delta_{i+1,j+1}) = \nu([P_M]_{j+1} + \delta_{i+1,j}) \\
\gamma_{i,j} \cdot (e_{i+1}, f_{j+1}) &
 \text{ otherwise}
\end{cases}
\end{align*}

\subsection{Correctness}

The next results 
\longversion{show}{state}
that the constructed alignment satisfies the requirements of our conformance checking task, cf. Def.~\ref{def:conformance_checking}.
\longversion{}{The formal proofs are omitted for reasons of space, but can be found in the extended version~\cite{BPMArxiv22}.
It is however easy to see that our encoding matches the same definitions as in Sec.~\ref{sec:preliminaries} and~\ref{sec:alignments}, and the cost functions in  Sec.~\ref{sec:costmodel}.}

\begin{lemma}
\label{lem:decoding}
For any satisfying assignment $\nu$ to \eqref{eq:constraints},
$(i)$ $\procrun_\nu$ is a process run, and $(ii)$ $\logtrace_\nu$ is a realization of $\ulogtrace$.
\end{lemma}
\longversion{
\begin{proof}
$(i)$
Let $M_i$ be the marking such that $M_i(p)= \nu(\markvar_{i,p})$, for all $p\in P$,
and $\alpha_i$  the state variable assignment such that
$\alpha_i(v) = \nu(\datavar_{i,v})$, for all $v\in V$ and $0\,{\leq}\,i\,{\leq}\,n$.
For $\procrun_{\nu} = \tup{f_1, \dots, f_n}$, we show by induction on $i$ that the transition sequence
$\procrun_{\nu,i} = \tup{f_1, \dots, f_i}$ 
satisfies 
$\smash{(M_I,\alpha_0)\goto{\procrun_{\nu,i}} (M_i,\alpha_i)}$ for all $0\leq i \leq n$.
In the base case $i=0$, so $\procrun_{\nu,i}$ is empty. As $\nu$ satisfies 
$\varphi_{\mathit{init,fin}}$, it must be that $M_0=M_I$ and $\alpha_0$ is the initial assignment, so the claim trivially holds.
In the inductive step, we consider $\procrun_{\nu,i+1} = \tup{f_1, \dots, f_{i+1}}$ and
assume that $\procrun_{\nu,i}$ satisfies 
$\smash{(M_I,\alpha_0)\goto{\procrun_{\nu,i}} (M_i,\alpha_i)}$.
For the last transition firing $f_{i+1} = (t_j,\beta)$ there must be some $j$ such that $1\,{\leq}\,j\,{\leq}\,|T|$ and $\nu(\transvar_{i+1}) = j$,
by construction and requirement (a) above.
Since $\nu$ is a solution to \eqref{eq:constraints}, it satisfies $\varphi_{\mathit{enabled}}$ so that $t_j$ is enabled in $M_i$.
Moreover, as $\nu$ satisfies $\varphi_{\mathit{mark}}$ and $\varphi_{\mathit{data}}$, we have
$\smash{(M_i,\alpha_i)\goto{f_{i+1}} (M_{i+1},\alpha_{i+1})}$.
This concludes the induction proof.
For the case where $i=n$, we thus obtain 
$\smash{(M_I,\alpha_0)\goto{\procrun_{\nu,n}} (M_n,\alpha_n)}$, and
$\procrun_{\nu,n} = \procrun_{\nu}$.
Finally, given that $\nu$ satisfies $\varphi_{\mathit{final}}$, the last marking $M_n$ must be final and hence $\procrun_\nu \in \runsof{\NN}$.

$(ii)$
Let $\sset[k]\uevent$ be all events $\uevent \in \ulogtrace$ such that $\nu(\dropvar_{\uevent}) = \bot$. By requirement (d), all events in $\ulogtrace \setminus \sset[k]\uevent$ are uncertain.
By construction, $\logtrace_\nu=  \tup{e_1, \dots, e_k}$ is obtained from $\sset[k]\uevent$ by taking for each $\uevent_i$ the timestamp value $t_i = \nu(\tsvar_{\uevent_i})$ in a way such that $t_1 \leq t_2 \leq \dots \leq t_k$. For all $1\leq i \leq k$, $t_i$ is an admissible timestamp for $\uevent_i$ by requirement (g).
We have $\lab(e_i) = \nu(\actvar_{\uevent_i})$, which is admissible by requirement (e), and 
$\hat{\alpha}(e_i)(v) = \nu(\tdvar_{v,\uevent_i})$ for all $v\in V$, which is admissible by requirement (f).
Thus $\logtrace_\nu$ is a realization of $\ulogtrace$ according to Def.~\ref{def:realization}.
\qed
\end{proof}
}{}
This lemma shows that the decoding provides both a valid process run and a trace realization. Next we demonstrate that the decoded alignment is optimal.
\longversion{To this end, we assume for the sake of simplicity that the final marking is non-empty, and admits a silent transition to itself; however, this restriction could be avoided by encoding refinements.}{}

\begin{theorem}
\label{thm:correctness}
Let $\NN$ be a DPN, $\ulogtrace$ a log trace with uncertainty and $\nu$ a solution to \eqref{eq:constraints} as in Sec.~\ref{ssec:decoding}.
Then $\gamma_{m,n}$ is an optimal alignment for $\ulogtrace$ of cost $\mathfrak{K}(\gamma_{m,n},\ulogtrace) =\nu(\distvar_{m,n})$.
\end{theorem}
\longversion{
\begin{proof}
By Lem.~\ref{lem:decoding}, $\procrun_\nu\in \runsof{\NN}$ and 
$\logtrace_\nu$ is a realization of $\ulogtrace$.
Let $\uevent_1, \dots, \uevent_m$ be the sequence of events in $\ulogtrace$ ordered in a way such that $\nu(\tsvar_{\uevent_1}) \leq \dots \leq \nu(\tsvar_{\uevent_m})$.
Moreover, let $\ulogtrace_i$ be the subset of $\ulogtrace$ such that $\ulogtrace_i = \{\uevent_1, \dots, \uevent_i\}$ for all $0\leq i \leq m$.
Let moreover $\widehat \logtrace_i$ be the projection of $\logtrace_\nu$ to $\ulogtrace_i$, i.e., the prefix of $\logtrace_\nu$ such that for all 
events in $\widehat \logtrace_i$ the respective event with uncertainty is in
$\ulogtrace_i$.
This subtrace with uncertainty is needed to perform the induction proof below.
Using the observations in the proof of Lem.~\ref{lem:decoding} (b), 
it is easy to see that 
$\widehat \logtrace_i$ is a realization of $\ulogtrace_i$ for all $i$, $0\leq i \leq m$.
Note that the length of the sequence $\widehat \logtrace_i$ is smaller or equal to $i$.

Let $d_{i,j} = \nu(\distvar_{i,j})$, for all $i$, $j$ with $0 \leq i \leq m$ and $0 \leq j \leq n$.
We show now the following $(\star)$: $\gamma_{i,j}$ is an optimal alignment of $\widehat \logtrace_i$ and $\procrun_\nu|_j$ with cost $\mathfrak{K}(\gamma_{i,j},\ulogtrace_i) = d_{i,j}$, by induction on $(i,j)$.
In the following, we freely use the fact that $[P_=]$, $[P_L]$, and $[P_M]$ are correct encodings of $P_=$, $P_L$, and $P_M$ from Exa.~\ref{exa:6}, cf.~\cite{cocomot}.
\begin{compactitem}
\item[\emph{Base case.}]
If $i\,{=}\,j\,{=}\,0$, 
then $\ulogtrace_i = \emptyset$ and $\gamma_{i,j}$ is the empty sequence, which is the optimal alignment of an empty log trace and an empty process run.
We have $d_{i,j}\,{=}\,0$ by $(\varphi_\delta)$, and also $\mathfrak{K}(\gamma_{i,j},\ulogtrace_i)=0$.
\item[\emph{Step case 1.}]
If $i\,{=}\,0$ and $j\,{>}\,0$, then the only possibility to match the last transition $f_j$ of $\procrun_\nu|_{j}$ is a model step with $f_j$.
By the induction hypothesis, $\gamma_{0,j-1}$ is an optimal alignment of the empty trace and $\procrun_\nu|_{j-1}$ of cost $\mathfrak{K}(\gamma_{0,j-1},\emptyset) = d_{0,j-1}$.
Thus, also $\gamma_{0,j}=\gamma_{0,j-1}\cdot\tup{(\empty, f_{j}) }$ is optimal.
We have $d_{0,j} = d_{0,j-1} + \nu([P_M]_j)$ by $(\varphi_\delta)$,
and by the choice of our cost functions,
$\mathfrak{K}(\gamma_{0,j},\emptyset) = \mathfrak{K}(\gamma_{0,j-1},\emptyset) + P_M(f_j) = d_{0,j-1} + \nu([P_M]_j)$.
\item[\emph{Step case 2.}]
If $j\,{=}\,0$ and $i\,{>}\,0$, then according to $(\varphi_\delta)$ either
$(i)$ $d_{i,0} = \nu([P_L]_{i}\cdot [\theta_{\ulogtrace}]_i) + d_{i-1,0}$
and $\gamma_{i,0}=\gamma_{i-1,0}\cdot\tup{(e_{i}, \empty)}$,
or
$(ii)$ $d_{i,0} = \nu([\kappa_{\ulogtrace}]_i) + d_{i-1,0}$ and
$\gamma_{i,0}=\gamma_{i-1,0}$.
Let $\uevent_i$ be the event with uncertainty in $\ulogtrace$ that matches $e_i$,
and $p$ be such that $(\lab(e_i)\colon p) \in \ueventLabels(\uevent_i)$.
By the induction hypothesis, $\gamma_{i-1,0}$ is an optimal alignment of 
$\widehat \logtrace_{i-1}$ and the empty run with cost $\mathfrak{K}(\gamma_{i-1,0},\emptyset) = d_{i-1,0}$.
In case $(i)$, $\nu([P_L]_{i}\cdot [\theta_{\ulogtrace}]_i) = 3-\conf{\uevent_i}-p$,
and a similar case distinction as Eq.~\eqref{eq:kappaenc} but for $[\theta_{\ulogtrace}]_i$ ensures that $\nu(\dropvar_{\uevent_i}) = \bot$,
so that $\mathfrak{K}(\gamma_{i,0},\emptyset) = d_{i-1,0} + \kappa(e_i,\empty) \otimes \theta(e_i,\ulogtrace_i)$ as desired,
according to our choices for the cost function and realization cost from Sec.~\ref{sec:costmodel}.
If case $(ii)$ applies, we can assume that $\nu([\kappa_{\ulogtrace}]_i) < \infty$,
so by Eq.~\eqref{eq:kappaenc} we must have $\nu(\dropvar_{\uevent_i}) = \top$,
and
$d_{i,0} = d_{i-1,0} + \ueventConfidence(\uevent_i)$, by our choice for the realization cost.
Requirement (d) implies that $\ueventConfidence(\uevent_i) < 1$, so $\uevent_i$ is
uncertain. Therefore, $\widehat \logtrace_i = \widehat \logtrace_{i-1}$ is a realization of $\ulogtrace_i$ where $e_i$ is dropped, and
$\gamma_{i,0}=\gamma_{i-1,0}$ a valid alignment.
According to $(\varphi_\delta)$, $d_{i,0}$ is assigned the minimum of the values 
corresponding to cases $(i)$ and $(ii)$, so since $\gamma_{i-1,0}$ is optimal, also
$\gamma_{i,0}$ is optimal.
\item[\emph{Step case 3.}]
If $i,j>0$, then, since $\nu$ satisfies $(\varphi_\delta)$, we can distinguish four cases:
\begin{inparaenum}
\item[$(i)$] $d_{i,j} = \nu([P_L]_{i}\cdot [\theta_{\ulogtrace}]_i) + d_{i-1,j}$,
\item[$(ii)$] $d_{i,j} = \nu([\kappa_{\ulogtrace}]_i) + d_{i-1,j}$,
\item[$(iii)$] $d_{i,j} = \nu([P_M]_{j}) + d_{i,j-1}$, and finally,
\item[$(iv)$] $d_{i,j} = \nu(ite([P_=]_{i, j} = 0, [\theta_{\ulogtrace}]_i, [P_=]_{i, j} + [P_=]_{i, j}\cdot [\theta_{\ulogtrace}]_i)) + d_{i-1,j-1}$.
\end{inparaenum}
In cases $(i)-(iii)$, we reason similarly as for cases $(i)$ and $(ii)$ in the Step Case 2, and as in Step Case 1, respectively,
to show that $\gamma_{i,j}$ is an alignment of $\widehat \logtrace_i$ and $\procrun_\nu|_j$ with cost $\mathfrak{K}(\gamma_{i,j},\ulogtrace_i) = d_{i,j}$.
In case $(iv)$, by the induction hypothesis, $d_{i-1,j-1}$ is the cost of the optimal
alignment for $\widehat \logtrace_{i-1}$ and $\procrun_\nu|_{j-1}$.
By construction, $\gamma_{i,j} = \gamma_{i-1,j-1} \cdot (e_i,f_j)$.
A similar case distinction as Eq.~\eqref{eq:kappaenc} but for $[\theta_{\ulogtrace}]_i$ ensures that $\nu(\dropvar_{\uevent_i}) = \bot$, so $e_i$ is included in $\widehat \logtrace_i$, so $\gamma_{i,j}$ is a valid alignment for $\widehat \logtrace_i$.
By Eq. \eqref{eq:costsum}, we have
$\mathfrak{K}(\gamma_{i,j},\ulogtrace_i) = d_{i-1,j-1} + \kappa(e_i,f_i) \otimes \theta(e_i,\ulogtrace_i)$, and
by Eq. \eqref{eq:costtheta},
$\kappa(e_i,f_i) \otimes \theta(e_i,\ulogtrace_i) = \nu(ite([P_=]_{i, j} = 0, [\theta_{\ulogtrace}]_i, [P_=]_{i, j} + [P_=]_{i, j}\cdot [\theta_{\ulogtrace}]_i))$,
so $\mathfrak{K}(\gamma_{i,j},\ulogtrace_i)$ is the cost of $\gamma_{i,j}$.

According to $(\varphi_\delta)$, $d_{i,j}$ is assigned the minimum of the values 
corresponding to cases $(i)-(iv)$, so $\gamma_{i,j}$ is optimal, which concludes the induction proof.
\end{compactitem}
We can assume that an optimal alignment $\gamma$ exists
where the process run $\restr{\gamma}{M}$ has exactly length $n$. While Lem.~\ref{lem:bound} guarantees that there is some $\gamma$ such that $\restr{\gamma}{M} \leq n$, we can assume $\restr{\gamma}{M} \geq n$
if for all $\alpha$ the final marking $M_F$ admits a step $\smash{(M_F, \alpha) \goto{(t,\beta)} (M_F, \alpha)}$ with a silent transition $t$. Such transitions can always be added to the net $\NN$.
Thus, the claim of the theorem follows from case $i=m$, $j=n$ of $(\star)$, since
$\ulogtrace_m = \ulogtrace$ and therefore $\widehat\logtrace_m = \logtrace_\nu$.
\qed
\end{proof}}
{}

Moreover, as explained in Sec.~\ref{sec:costmodel} (after Def.~\ref{def:conformance_checking}), we can easily capture the additional task of computing the lower-bound on the optimal cost of alignments of realizations for a given trace with uncertainty, as considered in \cite{PegoraroUA21}. 
By taking advantage of the modularity of our framework, this simply amounts to 
set $\kappa_{\ulogtrace} = 0$ and $\kappa(e_i,f_i)\otimes \theta(e_i,\ulogtrace)= \kappa(e_i,f_i)$,
thus ignoring all confidence values specified in $\ulogtrace$. 
This allows us to freely select, without any penalty, the realization of $\ulogtrace$ that has the minimal alignment cost. The following lemma formalizes this property:

\begin{lemma}
For $\NN$, $\ulogtrace$ as above and $\gamma_{m,n}$ the alignment decoded from a satisfying assignment $\nu$ for $(\Phi)$ as in Sec.~\ref{ssec:decoding},
there is no realization $\logtrace$ of $\ulogtrace$ and alignment $\gamma$ for $\logtrace$
such that $\kappa(\gamma) < \kappa(\gamma_{m,n})$.
\end{lemma}

Note that in contrast to the approach in~\cite{PegoraroUA21}, our approach entirely avoids any explicit construction of realizations, which is a huge benefit for the overall performance. 

\subsection{Implementation}\label{sec:implem}

As a proof of concept, the uncertainty conformance checking approach described in this paper was implemented in {\tool} -- a Python command line tool that was originally designed for data-aware conformance checking without uncertainties~\cite{cocomot}. It uses \texttt{pm4py} (\url{https://pm4py.fit.fraunhofer.de/}) 
to perform parsing tasks, and the SMT solvers 
Yices 2 \cite{Dutertre14} and Z3 \cite{deMouraB08}.

The tool takes as input two files: a DPN in \texttt{.pnml} format and a log in \texttt{.xes}, specified using the XES extension for uncertain data described in~\cite{Pegoraro21}.
The command line option \texttt{-u} triggers the use of the uncertainty module,
and the tool outputs the optimal alignment as well as its cost.
Based on the the encoding in Sec. \ref{sec:encoding}, the tool employs the two cost functions mentioned in Ex. \ref{exa:6} to achieve two different tasks:
Using the first cost function that takes confidence values into account, the cost of the optimal alignment can be interpreted as an expectation value of the best alignment cost for all realizations (parameter $\texttt{-u fit}$).
Using the second cost function, a lower bound on the cost of the optimal alignment among all realizations is computed (parameter $\texttt{-u min}$).
More information on the tool usage, the format for specifying uncertain logs, execution options and further details, together with the source code, can be found on the tool website.\footnote{\url{https://github.com/bytekid/cocomot}}

Although the presented encoding shows that the overall theoretical complexity of our approach does not change with respect to the one reported in~\cite{cocomot} (that is,  the problem of finding the optimal alignment for logs with uncertainty is NP-complete), experimental evaluations are required so as to assess the feasibility of the encoding in practical scenarios. More specifically, we plan to enrich publicly available logs for multi-perspective conformance checking~\cite{MannhardtLRA16} with uncertainty information, as done in~\cite{PegoraroUA21}.


\section{Conclusions}
\label{sec:conclusion}

In this work we have proposed an extension of the foundational framework for alignment-based conformance checking of data-aware processes studied in~\cite{cocomot}, to support logs with different types of uncertainties in events, timestamps, activities and other attributes. 
To account for all possible combinations of uncertainties in a trace, we rely on a notion of realization to fix one of its possible \emph{certain} variants.
However, given that there are potentially infinitely many realizations, performing the conformance checking task on each of them is not feasible. 

To attack this problem, we considered a version the conformance checking task aimed at searching for the best alignment among all possible realizations. This has been achieved by introducing an involved cost model that incorporates traditional alignment-related penalties together with extra costs accounting for the selection of specific realizations.
Although these cost components are not fixed and can in fact be tailored to specific settings and assumptions, we have provided a concrete instantiation and its corresponding encoding. 

We have also shown that, thanks to the modularity of our conformance cost definition, we
can accommodate different conformance checking tasks for logs with uncertainty, including those studied in the literature~\cite{PegoraroUA21}. 

The theoretical underpinning of our approach is SMT solving. Our work is the first one to employ techniques based on \emph{satisfiability} of formulae 
modulo suitable logical theories for solving data-aware conformance checking  tasks with uncertainty, and to leverage well-established solvers to handle them.
The approach was implemented in the 
\tool tool that is freely available.


In future work, we plan to investigate further, more involved notions of uncertain logs, and conduct an experimental evaluation of our approach and implementation. To this end, instead of considering artificially generated logs, one first step is to compile a benchmark for data-aware conformance checking of uncertain logs, which is currently not available.

\smallskip\noindent\textbf{Acknowledgments.}
This research has been partially supported by the UNIBZ projects VERBA, MENS, WineID, SMART-APP and by the PRIN 2020 project PINPOINT.

\bibliographystyle{abbrv}
\bibliography{references}
\end{document}